\documentclass[conference]{IEEEtran}
\usepackage{times}

\usepackage[numbers]{natbib}
\usepackage{multicol}
\usepackage[bookmarks=true]{hyperref}
\usepackage{graphicx}
\usepackage{amsmath}
\usepackage{mathrsfs}
\usepackage{amssymb}
\usepackage{bm}

\begin{document}

\title{Critical Scenario Generation for Developing Trustworthy Autonomy}

\author{Wenhao Ding\\ Carnegie Mellon University}

\maketitle
\IEEEpeerreviewmaketitle

\section{Motivation and Problem Statement}
\label{sec:introduction}

Autonomous systems, such as self-driving vehicles, quadrupeds, and robot manipulators, are largely enabled by the rapid development of artificial intelligence.
However, such systems involve several trustworthy challenges such as safety, robustness, and generalization, due to their deployment in open-ended and real-time environments.
To evaluate and improve trustworthiness, simulations or so-called \textit{digital twins} are largely utilized for system development with low cost and high efficiency. One important thing in virtual simulations is \textit{scenarios} that consist of static and dynamic objects, specific tasks, and evaluation metrics.

\begin{figure}[t]
  \centering
  \includegraphics[width=0.48\textwidth]{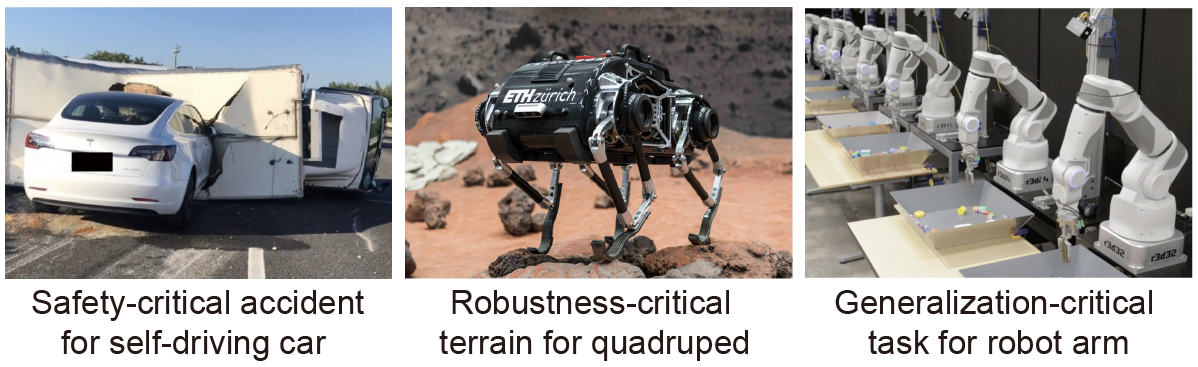}
  \vspace{-3mm}
  \caption{Examples of critical scenarios in autonomy.}
  \label{fig:example}
  \vspace{-6mm}
\end{figure}

\begin{figure*}[t]
  \centering
  \includegraphics[width=1.0\textwidth]{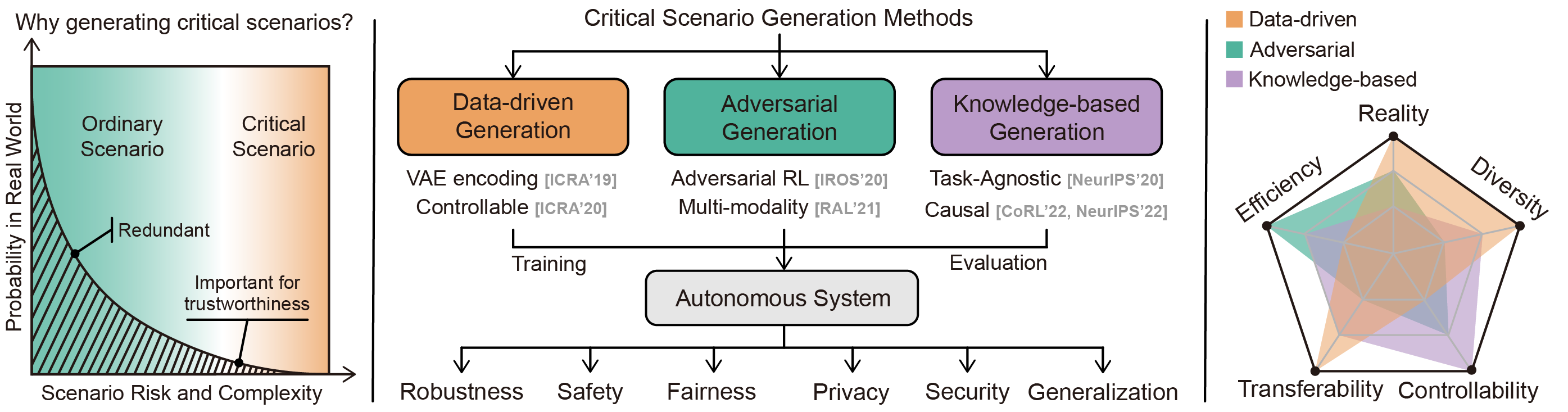}
  \caption{\textbf{Left:} Critical scenarios have a low probability to happen but are important for improving trustworthiness \textbf{Middle:} My contributed approaches can be divided into three categories. \textbf{Right:} Analysis of three types of methods from five perspectives.}
  \vspace{-1mm}
  \label{fig:overview}
  \vspace{-6mm}
\end{figure*}

However, designing diverse, realistic, and effective scenarios is still a challenging problem.
One straightforward way is creating scenarios through human design, which is time-consuming and limited by the experience of experts~\cite{carlachallenge}.
Another method commonly used in self-driving areas is log replay. 
This method collects scenario data in the real world and then replays it in simulations~\cite{scanlon2021waymo} or adds random perturbations~\cite{bansal2018chauffeurnet, tremblay2018training}. 
Although the replay scenarios are realistic, most of the collected scenarios are redundant since they are all ordinary scenarios that only consider a small portion of critical cases.
The desired scenarios should cover all cases in the real world, especially rare but critical events with extremely low probability. 
We provide three examples in Figure~\ref{fig:example}: safety-critical accident for self-driving vehicles~\cite{ding2020learning}, robustness-critical terrain for quadrupedal robots~\cite{made2023dreamwaq}, and generalization-critic tasks for robot arm~\cite{fang2021discovering}.
As shown in the left part of Figure~\ref{fig:overview}, ordinary scenarios have a high probability happen but are usually redundant for training learning-based algorithms. 
\textit{Critical scenarios} are rare but important to test autonomous systems under risky conditions and unpredictable perturbations, which reveal their trustworthiness~\cite{salem2015towards}.

There are 5 aspects we should consider for generating critical scenarios:
\begin{itemize}
    \item \textbf{Reality.} Since the ultimate goal is to deploy autonomous systems in the real world, we should ensure the scenarios have the chance to happen in real situations.
    \item \textbf{Efficiency.} Critical scenarios are usually rare in the real world. The generation needs to consider the efficiency and increase the density of critical scenarios.
    \item \textbf{Diversity.} Critical scenarios are diverse. The generation algorithm should be able to discover and generate as many different critical scenarios as possible.
    \item \textbf{Transferability.} Scenarios are dynamic due to the interaction between autonomous systems and their surrounding objects. The scenarios we generate should be adaptive to different autonomous systems.
    \item \textbf{Controllability.} We may want to reproduce or repeat one type of critical scenario rather than random ones. The generation method should be able to follow instructions or conditions to generate corresponding scenarios.
\end{itemize}

\textbf{My research objective is to generate critical scenarios in simulations to help evaluate and improve the trustworthiness of autonomy, including but not limited to safety, robustness, generalization, privacy, fairness, and security.} 
In particular, I leverage the huge success of deep generative models~\cite{roose2022brilliance, ramesh2022hierarchical} and the increased scalability of datasets in robotics areas, for example, self-driving~\cite{caesar2020nuscenes} and manipulation task~\cite{james2020rlbench}.
My research is built upon multiple machine-learning tools, including adversarial examples, reinforcement learning, and causal discovery.
Additionally, recent powerful simulation and digital twin platforms~\cite{dosovitskiy2017carla, makoviychuk2021isaac} almost bridge the gap between simulation and the real world, which enables the fidelity of scenarios to the greatest extent.

\vspace{-2mm}

\section{Contributed Approaches}
\label{sec:approach}

My previous research has contributed several important approaches toward the critical scenario generation area~\cite{ding2023survey}. These approaches fall into three categories according to the source of the information for generations, i.e., data-driven generation, adversarial generation, and knowledge-based generation (shown in the middle of Figure~\ref{fig:overview}).

\subsection{Data-driven Generation}

We first consider algorithms that only leverage information from datasets. 
For example, we control several vehicles to pass an intersection by making them follow the trajectories recorded in the same intersection in the real world.
We can use density estimation models $p_{\theta}(x)$ parameterized by $\theta$ to learn the distribution of scenarios, which enables the generation of unseen scenarios, where $x$ represents the general scenario.
I select Variational Auto-encoder (VAE)~\cite{kingma2013auto} as the estimation model, which has an encoder $q(z|x)$ and a decoder $p(x|z)$. The encoder projects the scenario data to a low-dimensional latent space, where we can randomly draw latent codes. These codes are then converted back to scenarios by using the decoder~\cite{ding2019new}. This framework provides a simple way to generate new scenarios but still has two drawbacks: (1) we do not know the critical level of generated scenarios (2) generated scenarios are not conditioned on road maps.
To solve these two problems, I propose a new framework~\cite{ding2020cmts} to learn the joint distribution of ordinary data and collision data. After training, we can do linear interpolation in the latent space to control the critical level of generation scenarios. The model also takes road maps as conditions to force scenarios to satisfy constraints.

\subsection{Adversarial Generation}

We then consider a more efficient way for generation, which actively creates critical scenarios by attacking autonomy.
For example, we control the steepness of the terrain to make the quadrupedal robot fail in locomotion tasks. Although robots learn to avoid falling into the ground, the scenario generation methods can gradually increase the difficulty to attack the robot.
This framework, named adversarial generation, is essentially a game between two components, one is the generator, and the other is the robot. 
We propose a Reinforcement Learning method in~\cite{ding2020learning} to maximize the collision rate between a self-driving vehicle and cyclists. This simple framework efficiently generates the position and velocity of cyclists to make self-driving vehicles fail. However, scenarios generated by this method usually lack diversity and reality due to the naive objective. 
In~\cite{ding2021multimodal}, we modify the adversarial generation framework with two improvements. We leverage a multi-modal flow-based model~\cite{dinh2016density} to represent the distribution of the scenario and also train a prior distribution of the real-world dataset as a constraint.

\subsection{Knowledge-based Generation}

Notice that scenarios are constructed in the physical world, therefore, need to satisfy physical laws and specific rules. 
The samples in the density we estimate or the adversarial examples we generate could easily violate these constraints. 
In addition, domain knowledge also improves the efficiency of the generation.
In my previous research, I consider knowledge injection from two perspectives.
The first one is designing a tree-structure representation of the scenario and back-propagating knowledge from leaf nodes to the root node~\cite{ding2021semantically}, which forces the generated scenarios to satisfy given constraints.
The second method is from a causality perspective, which aims to analyze scenarios and find the reasons that cause critical events.
In the first step, I assume the causality underlying the critical scenario is given as a graph. Then, I propose an autoregressive generation model to efficiently generate diverse critical scenarios guided by this causal graph~\cite{ding2022causalaf}.
Next, I investigate how to automatically discover causal graphs from an interventional dataset and use the discovered causality for better generation~\cite{ding2022generalizing}.

\section{Future Directions} 
\label{sec:conclusion}

\textbf{Develop new generation methods.}
In the right part of Figure~\ref{fig:overview}, I summarize the trade-offs of three types of generation methods. None of them is perfect to satisfy all requirements but the chance to combine the advantages of different categories still exists. In particular, large models and datasets provide powerful implicit neural models, while causal reasoning and knowledge injection enable explicit symbolic learning. The combined neural-symbolic system has great potential to improve current critical scenario generation methods.

\textbf{Joint learning of trustworthy agent and scenarios.}
The trustworthy problem cannot be solved only by developing a perfect agent. 
All robots have limitations and there always exist scenarios that cannot be solved. 
The ultimate goal of critical scenario generation is not to make the robot fail but to make it realize the boundary of feasible solutions.
To this end, we can focus on the joint learning of agents and scenarios, which leads to several directions such as automatic curriculum learning~\cite{narvekar2020curriculum} and adversarial training~\cite{shafahi2019adversarial}.

\textbf{Practical value of critical scenario generation.}
To push the academic research results to real-world applications, I am also working on a benchmark~\cite{xu2022safebench} that contains thousands of critical traffic scenarios for self-driving vehicles. This benchmark includes several important scenario-generation methods for evaluating object detection, trajectory prediction, and vehicle control algorithms. 
This benchmark will be extended to other areas where trustworthiness is also highly required as in self-driving.

\newpage
\bibliographystyle{plainnat}
\bibliography{arxiv}

\end{document}